\documentclass[conference]{ieeeconf}

\IEEEoverridecommandlockouts
\usepackage{amsmath,amssymb,amsfonts}
\usepackage{graphicx}
\usepackage{textcomp}
\usepackage{xcolor}
\usepackage{booktabs}
\usepackage{multirow}
\usepackage{tabularx}
\usepackage{ragged2e}
\makeatletter
\let\NAT@parse\undefined
\makeatother
\usepackage{hyperref}
\hypersetup{hidelinks}

\usepackage{caption}
\renewcommand{\thesection}{\arabic{section}}
\renewcommand{\thesubsection}{\thesection.\arabic{subsection}}
\renewcommand{\thesectiondis}{\thesection.}

\renewcommand{\thetable}{\arabic{table}}

\newcommand{\ssection}[1]{\noindent {\bf #1} }
\newcolumntype{L}{>{\RaggedRight\arraybackslash}X}

\begin{document}

\title{\LARGE \bf VREgo: Consumer VR Headsets for Learning Robot Manipulation\\from Egocentric Demonstrations}

% Authors: only the first author is filled in; add the co-authors and their affiliations.
\author{%
Liuchuan Yu$^{1}$%
% \thanks{$^{1}$George Mason University, Computer Science, Fairfax, Virginia 22030, USA. {\tt\small liuchuany@acm.org}}%
\thanks{$^{1}${\tt\small liuchuany@acm.org}}%
}

% \teaser{} (defined in ieeeconf.cls, as in the EgoKit paper) puts a full-width figure at the top of page 1.
\teaser{%
  \centering
  \includegraphics[width=\textwidth]{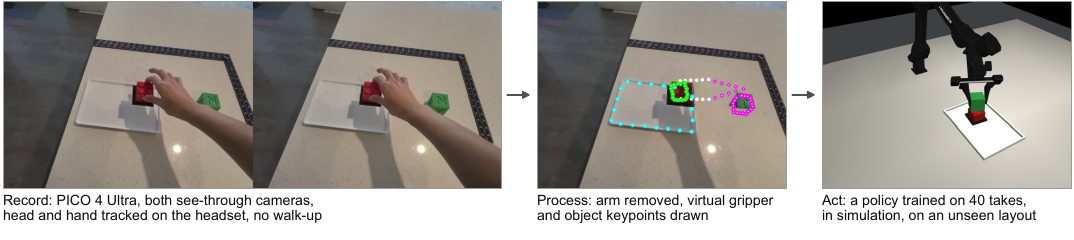}
  \captionof{figure}{\textbf{VREgo} trains robot manipulation policies from egocentric demonstrations recorded on consumer VR headsets, here the PICO 4 Ultra. Left: one frame of a two-move demonstration (the red block onto the tray, then the green cube onto the block's studs), as the headset's two see-through cameras recorded it; the headset tracked the wearer's head and hand on the device, and the wearer stood at the table from the first frame. Middle: the same frame as a training image, with the arm removed, a virtual gripper drawn at the tracked hand, and keypoints on the tray, the block and the cube. Right: in simulation, the baseline policy trained on 40 such demonstrations sets the cube on the block in an evaluation layout that no demonstration recorded. The baseline met both placement criteria in 2 of 100 evaluation layouts; after the changes of Section~\ref{sec:fix}, 93.8\,\% of episodes met them.}%
  \vspace{-3mm}
  \label{fig:teaser}%
}

\maketitle
\thispagestyle{empty}
\pagestyle{empty}

\begin{abstract}
VREgo trains manipulation policies from demonstrations recorded on consumer VR headsets; we instantiate it with EgoKit~\cite{yu2026egokit} on one such headset, the PICO 4 Ultra. On-device head and hand tracking and stereo object localisation allow the operator to begin a demonstration at the table. We adapt HumanEgo's preprocessing to estimate grasp thresholds per recording, update object positions after release, and propagate segmentation masks between keyframes. We evaluate policies in simulation on two tabletop tasks recorded by one operator. For cube placement, we compare PICO and Project Aria recordings of the same 41 layouts. With 40 training demonstrations, the PICO policies placed the cube on the tray in 67, 74, and 60 of 100 evaluation layouts across three training seeds. The corresponding Aria results were 73, 78, and 0. For a two-stage stacking task recorded with the PICO alone, the baseline placed the block on the tray in 94 of 100 layouts and met the end-of-episode criterion for both placements in two. In one diagnostic episode, repeated replanning deferred release beyond the eight executed steps. Demonstration replays also showed that the open gripper could displace the cube before grasping it. After adapting the training trajectories and adding task-specific execution rules that use simulator object positions, the system met both placement criteria in 93.8\,\% of episodes across four evaluation noise seeds. The same 100 layouts informed rule development and evaluation. VREgo is available at https://vrego.chuange.org/.
\end{abstract}

\section{Introduction}

Head-mounted cameras can record manipulation demonstrations without requiring an operator to teleoperate a robot. EgoMimic~\cite{kareer2025egomimic} co-trains on human and robot demonstrations, whereas EgoZero~\cite{liu2025egozero}, Phantom~\cite{lepert2025phantom} and HumanEgo~\cite{humanego2026} train manipulation policies from human demonstrations alone. We use HumanEgo as our reference. It represents a demonstration through hand trajectories, object keypoints and images in which a virtual gripper replaces the human arm. HumanEgo~\cite{humanego2026} reports training with approximately 30 minutes of human video per task.

EgoMimic~\cite{kareer2025egomimic}, EgoZero~\cite{liu2025egozero} and HumanEgo~\cite{humanego2026} use Project Aria glasses~\cite{engel2023project}. In our Aria workflow, Meta's Machine Perception Services (MPS) process uploaded recordings to estimate head and hand poses. HumanEgo~\cite{humanego2026} triangulates object keypoints from successive RGB views during an initial scene sweep, which can be a lateral head movement or a forward approach. We used a forward approach (the walk-up) for the matched-device comparison.

Consumer VR headsets offer a different profile: they track the head and both hands on the device, and several give applications access to their see-through cameras. The PICO 4 Ultra, which we use, has two such cameras 64\,mm apart, and EgoKit~\cite{yu2026egokit} records both cameras' images together with the head pose and 26-joint hand tracking. We present VREgo, an approach that learns robot manipulation policies from such recordings. It represents a demonstration as HumanEgo does, by the hand's trajectory, the objects' keypoints and an image without the arm, and learns the same kind of policy, so HumanEgo serves as the reference throughout. VREgo differs from it in four ways:
\begin{enumerate}
\item \textbf{Capture device.} A consumer VR headset, the PICO 4 Ultra, records the demonstrations and tracks the head and hands itself, and a stereo step locates the objects from its two cameras.
\item \textbf{Data collection.} Because stereo, not head motion, supplies the parallax, the wearer stands at the table from the first frame; there is no walk-up.
\item \textbf{Pipeline.} We adapted grasp labelling, object tracking and segmentation to a new operator and to recordings with two sequential moves.
\item \textbf{Execution.} For a two-move task, the training records are rewritten for a gripper, and rules for executing the policy's output guide release timing, grasp position and placement.
\end{enumerate}

We evaluate in simulation, using a MuJoCo arm, on two tasks recorded by one operator. Task~1, a cube onto a tray, was recorded on the same 41 layouts with the PICO and with Aria glasses, the reference's device. Section~\ref{sec:task1} summarizes the device comparison also reported in the unpublished revision of the EgoKit manuscript~\cite{egokitrevision}. Task~2, a block onto the tray and then the cube onto the block's studs, was recorded with the PICO alone and without a walk-up. We examine two failure modes of the baseline and evaluate trajectory and execution changes designed to address them.

Our contributions are:
\begin{itemize}
\item A capture path from a consumer VR headset, the PICO 4 Ultra, that reproduces all 864 training images in a format-conversion test on one Aria recording and which locates objects by stereo, so that a take needs neither a cloud service nor a walk-up;
\item A pipeline tuned to recordings from another device and operator, including two-move tasks, with each change measured on pilot takes, and fixes for defects in the reference implementation;
\item A simulated evaluation of closed-loop policy execution and replays every demonstration's own hand path before any training, as a check on the data;
\item A collection of 123 demonstrations on two tasks and the policies trained on them, an analysis of why the two-move task fails, and changes to the training records and to the execution of the policy's output that increase the proportion of episodes meeting both placement criteria from 2 to 93.8\,\% on the development layouts.
\end{itemize}

\section{Related Work}

\ssection{Robot learning from egocentric video.} EgoMimic~\cite{kareer2025egomimic} co-trains on Aria recordings of people and on robot data. EgoZero~\cite{liu2025egozero} learns from Aria recordings alone, with the hand and the objects reduced to 3D points. Phantom~\cite{lepert2025phantom} inpaints the human arm and overlays a rendered robot. HumanEgo~\cite{humanego2026} combines an inpainted image with tokens for the hand and the objects. EgoDex~\cite{egodex} collected 829 hours with Apple Vision Pro, using the headset's own hand tracking, and EgoVLA~\cite{yang2025egovla} and EgoVerse~\cite{punamiya2026egoverse} pretrain on or pool such data at scale. Our reference is HumanEgo, whose representation of a demonstration VREgo shares; VREgo differs in how demonstrations are captured, processed and executed.

\ssection{Capture devices.} Project Aria~\cite{engel2023project} records egocentric video that MPS processes; HOT3D~\cite{banerjee2024hot3d} includes recordings from Aria glasses and Quest~3 headsets, with motion-capture ground truth. UMI~\cite{chi2024umi} records with a handheld gripper and DexCap~\cite{wang2024dexcap} with motion-capture gloves. XR headsets are widely used to teleoperate robots, for example in Open-TeleVision~\cite{cheng2024opentelevision} and in XRoboToolkit~\cite{zhao2026xrobotoolkit}, which supports PICO headsets. EgoKit~\cite{yu2026egokit} exposes one recording workflow on six host devices, including the PICO 4 Ultra used here.

\section{VREgo}\label{sec:method}

\subsection{Representation and Policy}

VREgo starts from the released implementation of HumanEgo~\cite{humanego2026}, our reference, and keeps its representation of a demonstration and its policy; Section~\ref{sec:improvements} lists where the pipeline departs from it. Preprocessing turns a recording into per-frame images, camera poses and hand landmarks. In our processing configuration, motion-based phase labels separate the approach and stop from manipulation. The final second supplies training examples for the terminal-state output. A grasp label comes from the distance between the thumb and index fingertips; the configuration we evaluated counts the hand as closed while they are less than 10.5\,cm apart. Grounding DINO~\cite{liu2024groundingdino} and SAM~2~\cite{ravi2024sam2} detect and segment each object and the arm. Twenty keypoints are picked on each object's outline, tracked with CoTracker3~\cite{karaev2024cotracker3}, and triangulated during an \emph{object window}, which in the reference is the end of the approach. The arm is removed by inpainting~\cite{suvorov2022lama}, a virtual gripper is drawn at the hand, and the objects' keypoints are drawn on them. A carried object is attached to the hand while the grasp label is closed. A flow-matching policy~\cite{lipman2023flow} takes the image together with tokens for each hand (its pose and current grasp state) and each object, and predicts the hand's pose and grasp state over the next 50 frames. A robot gripper follows the prediction in a closed loop.

\subsection{Hardware}\label{sec:hardware}

\begin{table}[t]
\centering
\caption{The two recording devices: the reference's Project Aria glasses and VREgo's PICO 4 Ultra. Sizes are for the 41 takes of task~1, trimmed to the task.}
\label{tab:devices}
\footnotesize
\begin{tabularx}{\columnwidth}{@{}lLL@{}}
\toprule
 & \textbf{Project Aria Gen~1} & \textbf{PICO 4 Ultra (VREgo)} \\
\midrule
Cameras used & one RGB, fisheye, 1408\,$\times$\,1408 & two see-through, 1280\,$\times$\,960 each, rectified, 64\,mm apart \\
Image for the pipeline & undistorted, 640\,$\times$\,480, 90$^{\circ}$ wide & left eye, 640\,$\times$\,480, 76$^{\circ}$ wide \\
Frame rate & 30\,fps & 30\,fps \\
Head pose & MPS SLAM, in the cloud & headset runtime, on the device \\
Hand & MPS, 21 landmarks, in the cloud & OpenXR, 26 joints, on the device \\
Objects located by & head motion, 2\,s of the walk-up & stereo, 1\,s; walk-up optional \\
Preparation & upload and MPS processing & local recording ready \\
Data, 41 takes & 15\,GB (VRS and MPS) & 1.3\,GB (video and poses) \\
\bottomrule
\end{tabularx}
\end{table}

Table~\ref{tab:devices} compares the two devices. The Aria glasses recorded with HumanEgo's settings, and each take went through MPS for SLAM and hand tracking. On the PICO 4 Ultra, EgoKit recorded both see-through cameras side by side (2560\,$\times$\,960 at 30\,fps, 76$^{\circ}$\,$\times$\,61$^{\circ}$ per eye), together with the head pose, both hands' 26 OpenXR joints, and both cameras' intrinsics and extrinsics. The hand skeleton that EgoKit can paint into the video was switched off. A take is ready for processing when the recording stops.

A device-specific import stage converts the recording into images, camera poses and hand landmarks. Both devices then use the same representation, with device-specific settings for object localisation. For the PICO, this stage reads an EgoKit episode. The left eye is resized to 640\,$\times$\,480 (focal length 407\,px). The head pose is composed with the camera's extrinsic and converted from OpenXR's y-up frame to the pipeline's z-up frame. The 26 OpenXR joints are mapped to the 21 landmarks of MPS, and the wrist and palm normals, which MPS provides and OpenXR does not, are reconstructed from the knuckles. To test agreement between the two import paths, we rewrote a recording from HumanEgo's release as an EgoKit episode and processed it through VREgo's PICO path. The camera poses, the hand frames that serve as training targets, the grasp labels, the phases and the object poses match those from the Aria path to the reported precision of 0.00\,mm and 0.00$^{\circ}$, and all 864 training frames are identical. Only the reconstructed palm orientation differs (median 0.61$^{\circ}$), and it feeds no training target.

\subsection{Stereo Localisation of Object Keypoints}\label{sec:stereo}

With one camera, a keypoint can be triangulated only from views taken at different places, so the reference needs the head to move during the object window; on the Aria, the walk-up provides that. VREgo adds the other eye's view of the same instant to every view the triangulator receives. The pair is rectified, so each keypoint is matched by normalized cross-correlation of the object's pixels along the same image row, at the native resolution. The triangulator itself is unchanged. On static background points 0.5--0.6\,m away, triangulated twice from two halves of a 60-frame window while the wearer stood still, the two results disagreed by 1.1--4.3\,mm with the second eye and by 9--925\,mm from the head's motion alone.

For locally uniform edges parallel to an epipolar line, multiple candidate patches can yield similar correlation scores. The matcher rejects these ambiguous correspondences. On the tray this is every keypoint on its two long edges, 12 of 20. Our first version fell back to triangulation from head motion, which placed them up to 14\,cm too deep and moved the tray 1.3--3.4\,cm away from the camera, differently in every take. Such keypoints are now placed where their line of sight meets the plane of the object's stereo-located keypoints, if those lie within 1\,cm of a plane, or the straight edge between two located neighbours. Head motion is used only if the head moved 15\,cm, and otherwise the keypoint is dropped. Of the 4{,}100 keypoints of the 82 PICO takes, 70\,\% were located by stereo, 22\,\% on a plane and 8\,\% on an edge; 11 were dropped, and none was located from head motion.

\subsection{Recording without the Walk-Up}\label{sec:protocol}

\begin{figure*}[t]
  \centering
  \includegraphics[width=\textwidth]{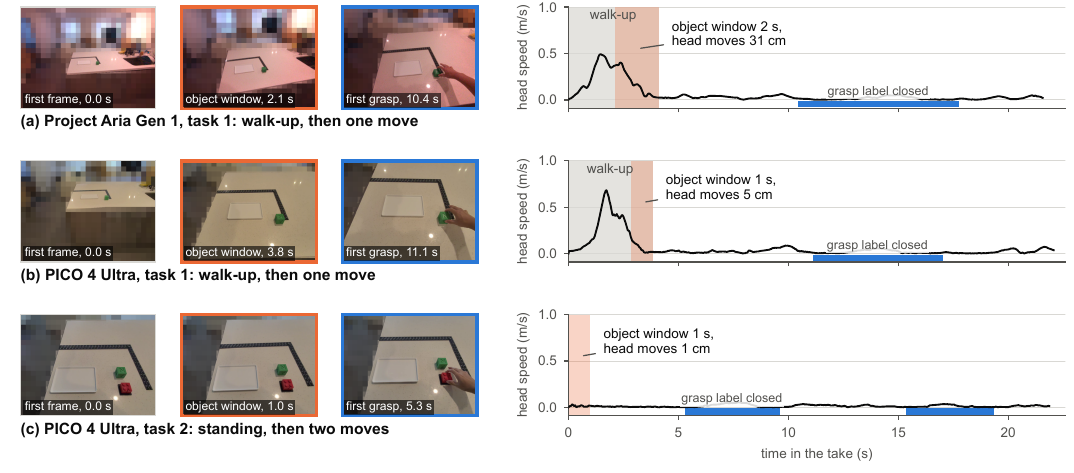}
  \caption{One take of each recording protocol: the first frame, the reference frame of the object window (orange border) and the first grasp (blue border), and the head's speed over the take. (a, b) Task~1 begins with a walk-up (gray), seen from across the room in the first frame. The object window (orange) is the last 2\,s of the walk-up on the Aria and its last second on the PICO. (c) Task~2 begins with the wearer at the table: the object window is the take's first second, in which the head barely moves, and the stereo pair supplies the parallax. Blue bars mark where the grasp label is closed: once in task~1, once per move in task~2. Over all 41 takes of each set, the head moved a median 36.5\,cm (Aria) and 6.0\,cm (PICO, task~1) between the first and last frames of the object window, against 0.8\,cm in task~2. Everything but the countertop is pixelated.}
  \label{fig:protocol}
\end{figure*}

Stereo provides a baseline for triangulation while the wearer stands at the table. Task~2 was therefore recorded without it (Fig.~\ref{fig:protocol}). The wearer starts the recording at the floor mark, keeps the hand down for about two seconds while looking at the objects, and then does the task. A task configuration flag selects this protocol. The object window becomes the take's first second, with its last frame as the reference frame. Those frames are taken out of the manipulation frames so that no stage sees them twice, and the checks that require an approach are dropped. At import, a take without a walk-up is kept from 1.5\,s before the hand first enters the image.

In the object window of task~2's takes the head moved a median 0.8\,cm (0.3--1.9\,cm), against 6.0\,cm in the last second of the PICO's walk-up in task~1 and 36.5\,cm in the Aria's two seconds. The walk-up itself took a median 3.8\,s and about a metre of head travel per take on either device. The protocol required the objects to remain visible throughout the approach. In three of our first seven pilot takes the wearer looked at the floor, and the objects were in view for the last second only.

\subsection{Improvements to the Pipeline}\label{sec:improvements}

\begin{table*}[t]
\centering
\caption{Where VREgo departs from the reference pipeline (HumanEgo), why, and what it changed. ``Pilot'' refers to the 14 takes recorded before the experiments' data (Section~\ref{sec:setup}), ``bulk'' to the 123 takes of the experiments.}
\label{tab:changes}
\footnotesize
\begin{tabularx}{\textwidth}{@{}p{0.2\textwidth}LL@{}}
\toprule
\textbf{Change} & \textbf{Why it was needed} & \textbf{Measured effect} \\
\midrule
Objects located by stereo, with keypoints on long edges placed on a plane or an edge (Section~\ref{sec:stereo}) & One camera needs head motion; an edge along the stereo baseline cannot be matched & Static points repeat to 1.1--4.3\,mm (9--925\,mm from head motion); no keypoint of the 82 PICO takes located from head motion \\
Stationary start (Section~\ref{sec:protocol}) & Monocular triangulation requires viewpoint displacement & Head moves 0.8\,cm in the object window; task~2's separation error $+$0.22\,$\pm$\,0.65\,cm (task~1: $+$0.25\,$\pm$\,0.49) \\
Grasp threshold per take; hand open until it is in the image & The reference's hand is closed below 10.5\,cm, set by its operators, whose open hand reads 14\,cm; ours read 7.6--9.6\,cm on the PICO, and a hand tracked before it enters the image hangs relaxed and reads closed & Label closes once per move in 11 of 11 pilot takes (7 of 9 with a fixed threshold) and in all 123 bulk takes \\
Objects drawn from their record & An object waiting for its turn loses its point tracks near the image border or under the hand, and they are not recovered & Pilot task~2: the cube's drawn keypoints off the cube in 51--80\,\% of frames before, 7--43\,\% after; bulk task~2: median 0\,\% \\
Carried objects located again after each release, in the data and in the simulator & A carried object is put down where the hand's record says, which depends on the label's timing; 12 pilot holds ended 0.3--2.8\,cm (median 1.5) from where the camera saw them & On two pilot takes, 0.1--0.5\,cm from the point tracks after release (2.3--2.7 before); bulk task~2: median 0.3\,cm (block) and 0.2\,cm (cube) from the camera's view \\
One detection box per object; isolated keypoints dropped; label runs under 20 frames removed & All detections above the threshold are merged into one mask, and the cube's prompt also boxed the block next to it; two stray keypoints moved a cube by 4\,cm & The cube located in the pilot take where it had not been; the stray keypoints removed \\
Segmentation of the video: SAM~2 carries masks between keyframes (every 15th frame, and every frame of the object window) & Per-frame detection and segmentation take 60--77\,\% of the processing time & Segmentation in a quarter to a third of the time; objects, gates and replays unchanged on 13 bulk takes (Table~\ref{tab:seg}) \\
Fixes to the released code & Batch processing drops every recording after the first; the grasp output is decoded at the value that means ``open''; the inference template crashes under NumPy 2.4; MPS hand poses are paired with images 1--2 frames late & Grasp decoding agrees with the held-out labels in 94.2\,\% of steps instead of 80.5\,\%; hands paired by timestamp ($\le$0.16\,ms); the late pairing would have affected 40 of our 41 Aria takes \\
\bottomrule
\end{tabularx}
\end{table*}

Table~\ref{tab:changes} lists what we changed. The pipeline was replicated before anything was changed: reprocessing two released recordings reproduced the authors' hand frames, phases and training targets exactly and their object positions within 1\,mm, and the released policy predicted the held-out recording as accurately from our preprocessing as from theirs (12.6 and 13.1\,mm). Each change to the pipeline is an option in the task file, off unless the task turns it on, so the released recordings still reproduce the authors' results. We describe the changes that matter most.

\ssection{Grasp label per take.} The reference configuration uses thresholds calibrated for its operators. Our operator's open hand read 7.6--9.6\,cm on the PICO, and a hold read 2.1--6.5\,cm, differently in every take, so any fixed threshold labelled some takes closed with nothing in the hand. VREgo splits each take's fingertip distance into its holding and open states, where the split explains most of the variance~\cite{otsu1979threshold}, and puts the threshold halfway between them. We assign the open label until the fingertips project into the image because the headset tracks the hand up to 2.4\,s before the camera sees it, hanging relaxed, and a relaxed hand reads as closed. In the main dataset, per-recording thresholds ranged from 5.9--8.4\,cm on the PICO and 8.3--9.0\,cm on the Aria, whose hand tracking reads a hold at about 6\,cm and an open hand at about 12.

\ssection{Carried objects.} In the reference, an object moves only while it is attached to the hand, so it is recorded to be put down where the hand was when the label opened. In our two-move pilot takes this put the carried objects 0.3--2.8\,cm from where the camera saw them afterwards, and the second move is aimed at the block. Neither attaching the object to another point of the hand nor shifting the label by a fixed rule brought the median below 1.0\,cm. VREgo therefore locates the object again after every release. In the frames that show it with no part of the arm on it, the ray through the centre of its mask is intersected with the table plane at the record's height, less the offset measured before the object was first touched, and the record is moved there. After release, the simulator obtains updated keypoints directly from the simulated object. A physical system would require an object-tracking or re-detection stage. The same takes showed why objects should be drawn from their record: the cube waiting for its turn lost its point tracks when the hand passed over it.

\ssection{Segmentation of the video.} The reference runs Grounding DINO and SAM~2 on every frame, for every prompt, with no memory of the previous frame, and this is most of the processing time. VREgo detects on keyframes and lets SAM~2's video predictor carry the masks in between. Every frame of the object window is a keyframe. Our first version carried masks there too, and they came out slightly larger; because the stereo step matches only the object's own pixels, edge keypoints then matched elsewhere, and objects moved by 1.4--11\,cm in 8 of 32 PICO takes. We detect the arm independently in each frame to prevent a false detection, such as a region of the white countertop, from propagating. Table~\ref{tab:seg} compares this with the per-frame default and with SAM~3.1~\cite{carion2025sam3,sam31release}, which we tested and did not adopt. SAM~3.1 removes most false arms, but its arm masks disagree with the default's in 9 of 38 takes, and in the five we inspected it had taken the arm's shadow for a second arm or dropped the forearm. These discrepancies escaped the quality checks and replays but could affect the images used for inpainting.

\begin{table}[t]
\centering
\caption{Segmentation backends on bulk takes. The per-frame backend is the reference's; VREgo uses SAM~2 as video.}
\label{tab:seg}
\footnotesize
\begin{tabularx}{\columnwidth}{@{}Lccc@{}}
\toprule
 & \textbf{Per frame} & \textbf{SAM~2 video} & \textbf{SAM~3.1} \\
\midrule
Segmentation, task-2 take 000 (s) & 455 & 114 & 488 \\
Objects, gates, replays & reference & same$^{1}$ & differ$^{2}$ \\
False-arm frames, 38 takes & 1{,}745 & 1{,}885$^{3}$ & 108 \\
Arm masks unlike the default's$^{4}$ & -- & none & 9 of 38 \\
\bottomrule
\end{tabularx}
\\[2pt]
\begin{flushleft}\footnotesize
$^{1}$On 13 takes that include every gate failure and every replay that changed in an earlier test: all estimated object positions agreed with the default to the reported precision (0.1\,cm), every gate and replay the same.\\
$^{2}$Gates passed in 25 instead of 24 of 38 takes, replayed moves 35 instead of 31 of 58, objects located over 1\,cm elsewhere in 10 takes.\\
$^{3}$Before the object-window fix, which does not change the arm.\\
$^{4}$Takes whose arm masks overlap the default's with an IoU below 0.7.
\end{flushleft}
\end{table}

\ssection{Additional variants.} Four further changes were measured on the pilot takes and not adopted. Taking the grasp label from the object's own motion fixed two task-1 gates on the Aria and one replay on the PICO, but in task~2 the block's mask sometimes also covered the cube behind it, which the label read as motion, and the stacks fell from 5 to 3 of 6. Dense stereo~\cite{hirschmuller2008sgm} inside the masks located the objects no better than the matched keypoints. Placing the Aria's objects on MPS's own table plane made the replays worse or changed nothing. An image-based hand tracker run on both eyes read holds more plausibly than the headset, but put carried objects down no closer to where they were (1.4 against 1.2\,cm).

\subsection{Simulated Evaluation and Checks before Training}\label{sec:sim}

The policy runs in a closed loop with a controller, which HumanEgo's release provides as a template for a real dual-arm robot. For the baseline, we retain the policy and controller, adapt two inference defaults described below, and implement the template's camera, arm and perception interfaces on MuJoCo~\cite{todorov2012mujoco}, with a Trossen WidowX AI arm from MuJoCo Menagerie~\cite{zakka2022menagerie}. The simulator renders the camera image from the wearer's viewpoint, takes the object masks from the scene, and removes the arm by rendering the scene without it. Two of the template's defaults had to be corrected: the grasp decoding of Table~\ref{tab:changes}, and the latching of the reference object, which training never does. With them, the reference's released policy set the loaf down on the plate in 47 of the 61 layouts of its own task (77\,\%), which checks the simulated loop against the reference; with the template's grasp threshold it did so in none of 21. For task~2, Section~\ref{sec:fix} also changes how the loop executes the policy's output.

The gripper is aligned with the recorded hand by a few values set per device from that device's recordings, never from how a policy fared: how far the gripper is pitched relative to the hand frame (15$^{\circ}$ for the PICO; 0$^{\circ}$ for the Aria, whose hand frame is pitched further down), which point of the gripper follows the pinch point, and how far the simulated masks are eroded. The jaws are kept level. In our first evaluation the simulator reported the levelled pose back to the policy, so the gripper drawn into its image lacked the 20--35$^{\circ}$ roll of the training images, and no policy grasped anything. All reported evaluations therefore feed the commanded pose back to the policy.

Each policy runs on the 41 recorded layouts of its task, with the scene rebuilt from each take, and on 100 shared layouts that no take recorded, drawn from the same ranges with another seed. For the shared layouts, the camera is placed at the median of the device's training-take camera poses relative to the layout's origin.

Before training, every take is checked in two ways. \emph{Quality gates} test, among other things, that the objects are located plausibly and stay in the image, that the hand is tracked in at least 95\,\% of the frames between its entry and the release, that the label closes once per move, that the drawn keypoints sit on their object in at least 90\,\% of the frames, and that a carried object's record ends within 3\,cm of where the camera sees it. A \emph{replay} makes the simulated arm follow the take's own hand path and grasp label in the scene rebuilt from it. In task~2 each move is aimed at where its target is in the simulation, because a policy reads where the objects are rather than repeating a path. Replay measures how reliably the recorded trajectories transfer to the simulated gripper.

\section{Experimental Setup}\label{sec:setup}

\begin{figure*}[t]
  \centering
  \includegraphics[width=\textwidth]{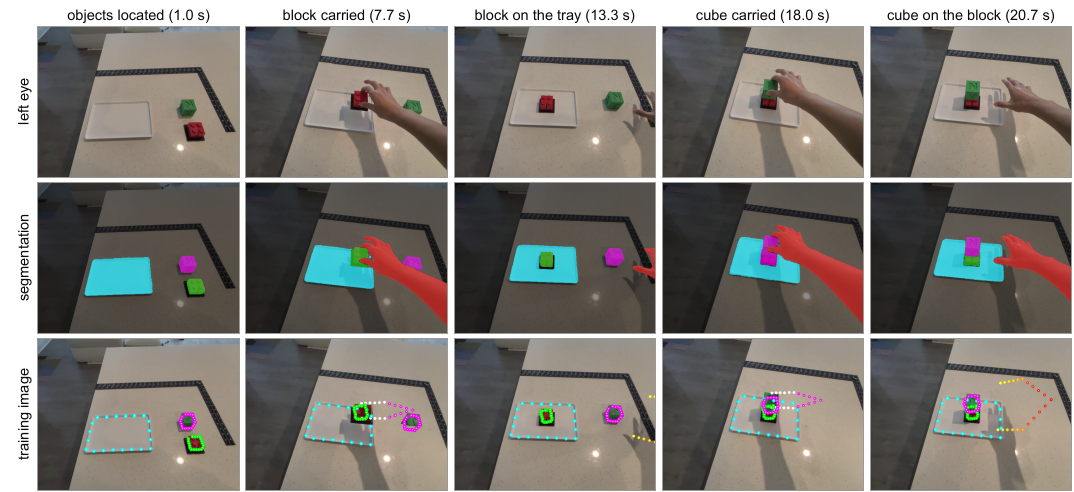}
  \caption{A task-2 take (layout 16) through VREgo's preprocessing, at five moments. Top: the left eye as the pipeline uses it. Middle: the detection masks of the arm (red), the tray (cyan), the block (green) and the cube (magenta). Bottom: the training image, with the arm removed, the virtual gripper drawn at the tracked hand (closed: white to magenta; open: yellow to red), and each object's keypoints drawn where its record has it. The block's keypoints follow it onto the tray, and the cube's onto the block, because each carried object was located again by the camera after its release. The arm's shadow is not part of the arm mask, so it stays in the training images.}
  \label{fig:task2data}
\end{figure*}

\ssection{Tasks.} Both tasks use a 5\,cm green 3D-printed cube and a white tray (26\,$\times$\,18\,cm, rim 1.2\,cm). Task~1 puts the cube on the tray, with $\pm$10\,cm of room sideways and $\pm$5.7\,cm in depth. Task~2 adds a red LEGO block (4.7\,$\times$\,4.7\,$\times$\,3.3\,cm on a 6.3\,cm base plate, with four studs whose centres are 2.5\,cm apart): the block goes onto the tray, and then the cube onto the block's studs. The second move must land within about $\pm$2\,cm: in simulation, a cube set down off-centre stays on the studs up to 1.6\,cm along the block's sides and 2.2\,cm along its diagonal. Every object stands square to the table edge, because the pipeline locates positions and gives every object fixed upright axes.

\ssection{Layouts and recording.} For each task, 41 layouts were drawn with a fixed seed and marked on a sheet taped to a white kitchen counter. In task~1 the cube lies anywhere in a 16\,$\times$\,16\,cm zone and the tray 22--32\,cm from it, within 25$^{\circ}$ of straight left. In task~2 the block lies in a 12\,$\times$\,12\,cm zone, the cube 10--14\,cm behind it, and the tray 24--30\,cm from the block. One operator recorded every layout. Task~1 was recorded once per layout with each device, with a walk-up, in blocks of five layouts alternating between the devices. Task~2 was recorded with the PICO only, without a walk-up, in two sittings of about an hour. Inspection of the raw streams identified three takes for re-recording (two Aria takes of task~1, and a task-2 take whose tray stood with its coordinates swapped) and six to trim. All takes were then cut to the task, with the poses kept aligned to the frames by timestamp: 14.6\,min (PICO) and 14.1\,min (Aria) for task~1, and 16.2\,min for task~2. Task~2 is PICO-only because in our pilot no Aria take could set the cube on the studs, even when replayed: at the grasp, MPS placed the hand 1--7\,cm from the cube along the camera's line of sight.

\ssection{Training and evaluation.} Policies were trained with the reference's recipe (400 epochs) on the first 10, 20 and 40 takes after take 000, which is held out. The 40-take policies of task~1 were trained with three seeds and all other conditions with one. Every take was used, including those that miss a gate. We count an object as \emph{set down} when it rests on its target as the policy's loop ends, after at most 110 policy steps in task~1 and 170 in task~2; task~2 is scored per move. The criterion permits the gripper to remain closed and is assessed before the simulator opens it during cleanup. The open-loop error is the mean distance between the predicted and the recorded hand position over the 50-frame horizon on the held-out take, compared with a hand that stays still. For task~1, the two devices' policies run on the same shared layouts, so we compare them pairwise: only layouts where exactly one of them succeeds count (two-sided exact sign test). The comparison's rules, including its primary outcome (set down on the shared layouts at 40 takes), were fixed before the takes were recorded. Preprocessing, training and evaluation ran on one NVIDIA H200 that another job shared. The policy samples every chunk from one noise tensor, drawn once per evaluation run from its seed, so a seed amounts to a different deterministic policy; the results of Section~\ref{sec:fix} are means over four seeds, summarized over the same 100 layouts at each seed. The diagnostic runs of Sections~\ref{sec:why} and~\ref{sec:fix} also used an RTX 3090, and comparisons are made on one GPU, because long episodes can differ between the two.

\section{Results}\label{sec:results}

\begin{table}[t]
\centering
\caption{Data and policies of the two tasks, 41 takes per column. Policies use seed~7.}
\label{tab:results}
\footnotesize
\setlength{\tabcolsep}{3.5pt}
\begin{tabularx}{\columnwidth}{@{}Lccc@{}}
\toprule
 & \multicolumn{2}{c}{\textbf{Task 1}} & \textbf{Task 2} \\
\cmidrule(lr){2-3}\cmidrule(l){4-4}
 & \textbf{Aria} & \textbf{PICO} & \textbf{PICO} \\
\midrule
\multicolumn{4}{@{}l}{\textit{Data}} \\
Walk-up & yes & yes & no \\
Take length, median (s) & 20.0 & 21.5 & 23.7 \\
Training frames per take, median & 363 & 377 & 616 \\
Preprocessing per take, median (min) & 9.0 & 8.3 & 12.8 \\
Takes passing all gates & 39 & 33 & 37 \\
Separation error, mean (SD) (cm)$^{1}$ & $+$0.08 (0.38) & $+$0.25 (0.49) & $+$0.22 (0.65) \\
Pinch offset, along / across (cm)$^{2}$ & $+$1.8 / 2.6 & $-$0.9 / 1.6 & $-$0.1 / 1.1 \\
Replay: object on its target & 33 & 29 & block 31, cube 8 \\
\midrule
\multicolumn{4}{@{}l}{\textit{Policies, 10 / 20 / 40 training takes}} \\
Open-loop error (mm) & 34 / 30 / 34 & 25 / 26 / 28 & 43 / 40 / 39 \\
\quad hand stays still (mm) & 37 & 47 & 51 \\
Set down, 100 unseen layouts & 59 / 66 / 73 & 35 / 70 / 67 & 0 / 3 / 2$^{3}$ \\
Set down, 41 recorded layouts & 17 / 31 / 36 & 16 / 29 / 33 & 0 / 1 / 3$^{3}$ \\
Training at 40 takes (h) & 5.3 & 5.2 & 8.8 \\
\bottomrule
\end{tabularx}
\\[2pt]
\begin{flushleft}\footnotesize
$^{1}$Distance between the located objects (task~2: tray and block) minus their distance on the layout sheet.\\
$^{2}$Median offset of the pinch point between the tracked thumb and index fingertips from the located object when the first grasp closes, along and across the camera's line of sight; negative is nearer the camera.\\
$^{3}$Both moves. The block alone: 28 / 71 / 94 (unseen) and 19 / 32 / 36 (recorded). With the changes of Section~\ref{sec:fix}: 93.8 of 100 unseen layouts at 40 takes.
\end{flushleft}
\end{table}

\begin{figure}[t]
  \centering
  \includegraphics[width=\columnwidth]{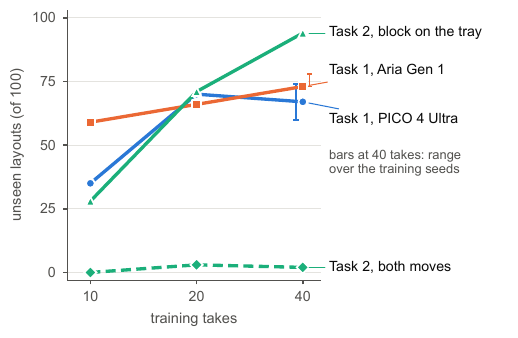}
  \caption{Layouts, of 100 that no take recorded, in which the policy set its object down, against the number of training takes. Lines are seed~7. The bars at 40 takes span task~1's three training seeds (for the Aria, two: its seed-9 policy declares the task finished at its first step and never moves).}
  \label{fig:results}
\end{figure}

\subsection{Task 1: The Devices Compared}\label{sec:task1}

Table~\ref{tab:results} and Fig.~\ref{fig:results} summarize task~1; Table~5 of the unpublished EgoKit revision~\cite{egokitrevision} gives the full device comparison. Mean signed inter-object separation errors were small (0.8\,mm for the Aria and 2.5\,mm for the PICO). When the grasp closes, the PICO's own hand tracking puts the pinch point nearer the located cube than MPS does, along the line of sight as well as across it. Fewer PICO takes pass every gate (33 against 39); in half of the PICO's misses the cube left its narrower view for part of the take. Replay placed the cube on the tray for 29 PICO takes and 33 Aria takes. The policies learned the task from both devices' data. With 20 or 40 takes they set the cube down in 60 to 78 of the 100 unseen layouts, except the Aria's seed-9 policy, which declares the task finished at its first step. With 10 takes the Aria policy was ahead (59 against 35, one seed each). With 40 takes, the PICO and Aria success counts were 67/73, 74/78 and 60/0 for training seeds 7, 8 and 9, respectively. For the first two seeds, the paired comparisons gave the following results: the PICO policy alone succeeded in 16 layouts and the Aria policy alone in 22 with seed~7 ($p=0.42$), and 12 against 16 with seed~8 ($p=0.57$). On the held-out take, the PICO policies' error was 25--28\,mm against 47\,mm for a hand that stays still, and the Aria policies' 30--34\,mm against 37\,mm.

On this task, PICO demonstrations supported successful simulated placements while requiring 1.3\,GB of storage, compared with 15\,GB for the Aria recordings and MPS outputs (Table~\ref{tab:devices}). The small number of training seeds limits conclusions about differences between devices.

\subsection{Task 2: Two Moves without a Walk-Up}\label{sec:task2}

\begin{figure*}[t]
  \centering
  \includegraphics[width=\textwidth]{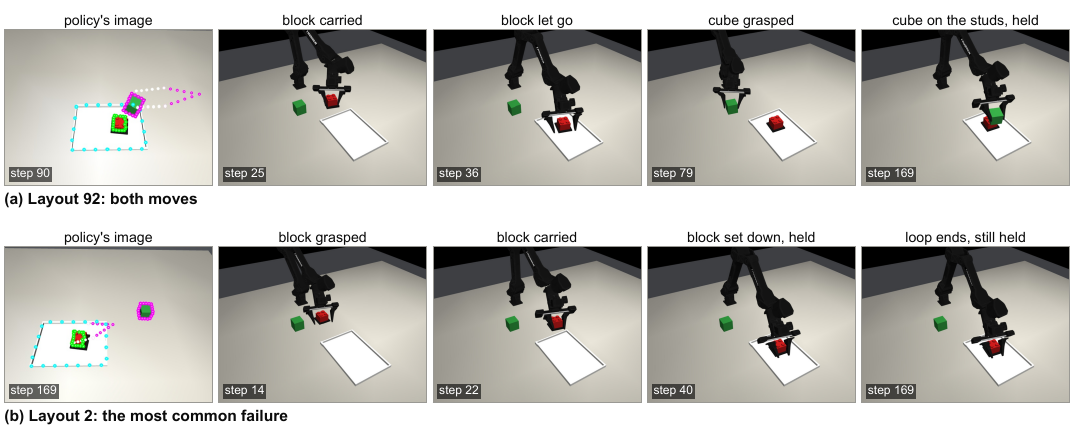}
  \caption{The 40-take task-2 policy before the changes of Section~\ref{sec:fix}, in simulation, on two of the 100 unseen layouts. The first column shows the policy input. The remaining columns show the episode from an external viewpoint. (a) One of the two layouts in which both moves succeed. The policy carries the block onto the tray, lets go, picks up the cube, and sets it on the studs 2.2\,cm from the block's centre, at the edge of what the studs allow. It never opens the gripper again; when the evaluation loop ends and its clean-up opens the gripper, the cube tips off. (b) The most common outcome: the block is set down on the tray and held until the loop ends, while the gripper does not reach the cube. The episodes were re-rendered on another GPU: (b) repeats the evaluated episode exactly, and (a) departs from it by 0.1\,mm at step 42 and ends the same way.}
  \label{fig:task2sim}
\end{figure*}

\ssection{The data.} The 41 takes of task~2 last a median 23.7\,s, from the wearer at the table to the hand withdrawn after the second move, and yield a median of 616 training frames per take, compared with 377 for PICO task~1 (Table~\ref{tab:results}). Fig.~\ref{fig:task2data} follows one take through the pipeline. Without a walk-up, the signed separation error between tray and block is $+$0.22\,cm on average, with an SD of 0.65\,cm. Every retained keypoint was located by stereo or placed on a plane or an edge, and none from head motion. In all 41 takes the hand was tracked in every frame between its entry and the first release, and the grasp label closes exactly twice, with per-take thresholds of 5.9--8.4\,cm. After their release, the block's and the cube's records end a median 0.3 and 0.2\,cm from where the camera sees them, and their drawn keypoints are on them in a median 100\,\% of the frames. Four takes miss a gate, all for the same reason: the cube was located implausibly, with keypoints spread over 1.8\,m. SAM~3.1's masks avoided this in one of them.

\ssection{Replay.} Replayed through the simulated arm, the takes put the block on the tray in 31 of 41 cases, but the cube on the block in only 8. Where the gripper released the cube at the height of the studs (24 takes), it did so a median 1.9\,cm from the block's centre (0.2--16.2\,cm). The 8 cubes that stayed were released within 1.9\,cm (median 1.2), and those that did not, a median 3.0\,cm off. The studs allow about 2\,cm. Under this replay protocol, the recorded trajectories transferred successfully to the simulated stacking task in approximately one fifth of the recorded layouts.

\begin{table}[t]
\centering
\caption{How the task-2 policies' episodes ended on the 100 unseen layouts, before the changes of Section~\ref{sec:fix}.}
\label{tab:task2}
\footnotesize
\begin{tabularx}{\columnwidth}{@{}Lccc@{}}
\toprule
\textbf{Training takes} & \textbf{10} & \textbf{20} & \textbf{40} \\
\midrule
Block not picked up & 69 & 29 & 6 \\
Block picked up, not set down on the tray & 3 & 1 & 0 \\
Block set down; gripper never opens again & 23 & 65 & 49 \\
Gripper opens, never closes again & 4 & 0 & 22 \\
Gripper closes again, cube not picked up & 0 & 1 & 11 \\
Cube picked up, not set down on the block & 1 & 1 & 10 \\
Cube set down on the block (both moves) & 0 & 3 & 2 \\
\bottomrule
\end{tabularx}
\end{table}

% AUTHOR CHECK: The 20-take first-placement count is reported as 71, whereas the mutually exclusive outcomes in the episode table imply 70. Verify against the logs and update the figure and table footnote together.
\ssection{Policies.} First-move placements increased from 28 to 94 of 100 evaluation layouts between the 10- and 40-take conditions. Both placements met the scoring criterion in 0, 3 and 2 layouts for the 10-, 20- and 40-take policies, respectively (Fig.~\ref{fig:results}). On the held-out take, open-loop error decreased from 43 to 39\,mm with more takes, against 51\,mm for a hand that stays still, and blanking the image changes the predictions by 5--10\,mm.

Table~\ref{tab:task2} shows how the episodes ended. The most common outcome at 20 and 40 takes is a block set down on the tray and held there until the loop ends: the policy never opens the gripper again and never turns to the cube (Fig.~\ref{fig:task2sim}b). With 40 takes it opened the gripper after the block in 45 of the 94 episodes in which the block was set down. Of those, 22 never closed it again, 11 closed it without picking up the cube, 10 picked the cube up but did not set it on the block, and 2 completed both moves.

\subsection{Why the Second Move Fails}\label{sec:why}

We probed the 40-take policy at every call of one unseen layout, repeating each call with altered inputs, and replayed the takes, logging where the cube was when the gripper closed. The policy samples from a fixed noise tensor, so a probe cannot change the episode. Two causes stand out, one in the closed loop and one in how the gripper meets the cube.

\ssection{Replanning defers release.} In task~1, opening the gripper is not required for success, and the policies seldom did it: the PICO's 40-take policy opened the gripper itself in 9 of the 67 episodes in which it set the cube down, and the Aria's in 24 of 73. In task~2 the second move cannot start without it. The predicted grasp follows the hand token's current grasp state: flipping that state flips the first predicted grasp step in 146 of the episode's 155 calls, while redrawing the gripper in the image or blanking the image changes nothing. Yet with the gripper closed, 144 of the 155 predicted 50-step chunks do contain the release, always at their 48th step. The loop executes 8 steps and predicts again; the hand stood still during those steps, so the next observation, and with it the next chunk, is the same, and the release never comes. In the demonstrations the label opens when the fingers spread, a median 28 frames (about 1\,s) after the object is set down, with the hand still, and a policy that sees only the present cannot count that pause. Human motion is bursty elsewhere too: between the block's release and the cube's grasp, 22.5\,\% of the open-gripper frames hold still for 8 frames and then move over 4\,cm, and the same fixed point later stopped retrained policies between the moves. Wen et al.~\cite{wen2020copycat} describe action copying in policies conditioned on observation histories. Here the policy observes the current grasp state. In the probed episode, its predictions preserved that state throughout each executed prefix, repeatedly deferring the planned release.

\ssection{Contact before grasp closure.} The replays place the cube on the block in 8 of 41 takes, although the real cube stood on the real block at the end of every take. At grasp closure, the cube had already moved a median of 1.5\,cm. Displacement exceeded 1\,cm in 16 of the 22 grasps examined. The open jaws are 9\,cm wide, pitched down, and reach about 2.3\,cm below and ahead of the pinch point. Following the person's hand, which comes in low from the side, they hit the cube before the grasp. Centring the grasp alone yielded successful stacking in 7 of 41 replays, compared with 8 at baseline. In our pilot, the Aria's hand tracking was further off along the line of sight at the grasp, and no Aria take could stack the cube at all.

\subsection{Trajectory and Execution Changes for Stacking}\label{sec:fix}

\begin{figure}[t]
  \centering
  \includegraphics[width=\columnwidth]{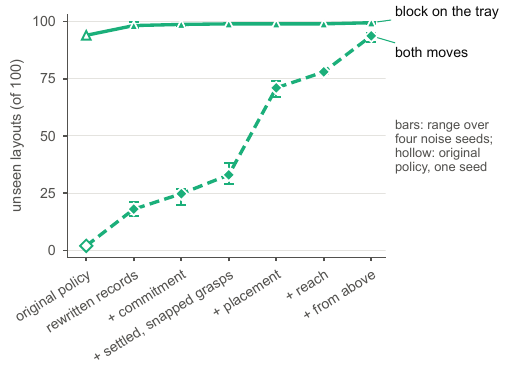}
  \caption{Number of development layouts (out of 100) in which the 40-take task-2 policy set the block on the tray and completed both moves: the original policy (hollow markers; one noise seed), then the policy retrained on the rewritten records, with the rules for executing its output added one at a time (Table~\ref{tab:fix}). Points are the mean over four noise seeds; bars span the seeds.}
  \label{fig:fix}
\end{figure}

We addressed deferred release by modifying training trajectories and execution timing, and pre-grasp contact by modifying the approach trajectory. We retrained the 40-take policy on rewritten records and added rules to the execution loop, keeping the model, the training recipe, the controller and the simulator unchanged, and evaluated each step on the same 100 development layouts with four noise seeds (Fig.~\ref{fig:fix} and Table~\ref{tab:fix}).

\ssection{Records rewritten for a gripper.} Each take's hand record is rewritten before training; the images are not. During each hold the hand is moved horizontally onto the held object's centre. In the last 30 frames before a grasp it rises to 6\,cm above the grasp point and comes straight down, and after a release it leaves straight up. Elsewhere the open hand stays 6\,cm above any object within 10\,cm. Replayed, records with these changes complete both moves in 20 of the 41 takes instead of 8. The clearance is then faded out by 14\,cm and limited to 5\,mm of change per frame, and a time warp keeps a frame only once the hand has moved 1.5\,mm from the last frame kept, or near a change of the grasp label, so that no chunk of training targets holds still and then moves; 74\,\% of the frames remain. Trained on these records with the same recipe, the policy completes both moves in 18.0 of 100 layouts with the loop unchanged, and it lets go of the block by itself: only 5.5 episodes hold it to the end, against 49 for the original policy.

\ssection{Rules for executing the output.} We then applied the following cumulative rules during execution, using the same retrained policy throughout.
\begin{itemize}
\item \emph{Commitment.} When the gripper is closed and a chunk plans the release, the chunk is executed through the release rather than for 8 steps.
\item \emph{Settled, snapped grasps.} A chunk that closes the gripper is shifted sideways so that the closing lands on the nearest object's estimated centre, if that is within 5\,cm, and the arm waits at the closing point until the tool arrives before it closes; the controller's smoothing leaves the tool about 1\,cm behind a descending target.
\item \emph{Placement.} While the cube is held, the chunk is shifted so that the cube is centred over the block at the bottom of the chunk's last descent, kept 3\,mm above resting on the studs, and released only once the tool is there. A closing higher than 3\,cm above the object's centre is lowered to that height.
\item \emph{Reach.} A closing that stops 5--12\,cm short of the next object to pick is extended to the object.
\item \emph{From above.} Every snapped approach to that object stays 3.5\,cm above its top until it is over it, then comes straight down and closes in the same call. When the tool has not moved since the previous call, 24 steps of the chunk are executed instead of 8.
\end{itemize}
The placement rule locates the held cube and the block by their keypoints on the simulated bodies, the re-detection the simulator already performs after a release, and the reach rule uses release information to identify the next object. The other rules use the policy's own object estimates.

\begin{table}[t]
\centering
\caption{The 40-take task-2 policy retrained on the rewritten records, with the rules of Section~\ref{sec:fix} added one at a time: outcomes on the 100 development layouts, averaged over four noise seeds. The original policy completed both moves in 2.}
\label{tab:fix}
\footnotesize
\setlength{\tabcolsep}{4pt}
\begin{tabularx}{\columnwidth}{@{}Lccccc@{}}
\toprule
\textbf{Execution} & \textbf{Both moves (SD)} & \textbf{(a)} & \textbf{(b)} & \textbf{(c)} & \textbf{(d)} \\
\midrule
Loop unchanged & 18.0 (2.4) & 5.5 & 49.3 & 8.0 & 17.5 \\
+ Commitment & 24.8 (3.3) & 0 & 12.5 & 44.5 & 17.0 \\
+ Settled, snapped grasps & 33.0 (3.7) & 0 & 9.0 & 14.8 & 42.3 \\
+ Placement & 71.0 (3.2) & 0 & 9.0 & 9.5 & 9.5 \\
+ Reach & 78.0 (0.8) & 0 & 6.5 & 6.5 & 8.0 \\
+ From above & \textbf{93.8} (1.9) & 0 & 2.5 & 2.3 & 1.0 \\
\bottomrule
\end{tabularx}
\\[2pt]
\begin{flushleft}\footnotesize
(a) Block set down; gripper never opens again. (b) Gripper opens, never closes again. (c) Gripper closes again, cube not picked up. (d) Cube picked up, not set down on the block. The block is picked up in every episode; it misses the tray in a mean of 0.5--1.75 episodes per 100, depending on the condition.
\end{flushleft}
\end{table}

Each cumulative change in Table~\ref{tab:fix} increased the mean two-move placement score on the development layouts. The four noise seeds repeat the same layouts, so we report these comparisons descriptively. Executing through the planned release eliminated episodes in which the gripper held the block until termination. Aligning the grasp and waiting for the tool reduced failed cube pickups, but many cubes were subsequently placed beside the block. Two things put them there. Chunks that carry the cube and release it in one call were not centred by an earlier version of the placement rule (64.5 both moves), because their lowest point before the release is the chunk's start, where the cube has just been lifted. And the policy plans the release 1.5--2.2\,cm lower than the studs let the tool go, so the cube was pressed onto them, tilted, and tipped off once released. Most of the failures that remained were grasps. Closings stopped a median 6.7\,cm short of the cube, after which the policy carried an empty gripper to the block, because a closing within 20\,cm attaches the nearest object to the hand. Taken on to the cube at the policy's low approach height, such closings still failed: the fingers rode up the cube's side. Approached from above, they hold. With all rules, both moves succeed in 93.8 of 100 layouts (91, 95, 95 and 94 over the four seeds); against the 64.5 with the earlier placement rule, 118 paired layout--seed episodes changed from failure to success and one changed from success to failure. What remains, per 100 episodes, is 2.5 arms that stop 16--23\,cm short of the cube near the arm's base, 2.3 failed cube grasps, 1.0 cube that does not stay on the block, and 0.5 blocks not set down.

\ssection{Additional variants.} Moving each release to the hand's rest emptied the pause before it but left the person's rest after it, and the policy then hovered after letting go. Two further rules, skipping a planned pause and opening the gripper at the arm's lowest permitted height, are not needed once the records are time-warped: the second never fires, and the observed change from the first was small. Correcting the height of every released object's record made the policy stop short of the cube more often (70 against 77 on one seed): the records of objects that were really carried are best relocated as in the training data, on the table plane.

\subsection{Cost}

The 41 takes of task~2 were recorded in two sittings of about an hour. Neither PICO task needed anything between recording and processing, whereas every Aria take had to be uploaded and processed by MPS, and the Aria's 41 task-1 takes occupy 15\,GB against the PICO's 1.3\,GB. Preprocessing took a median 8.3\,min per PICO take of task~1, 9.0\,min per Aria take, and 12.8\,min per task-2 take, three takes at a time on one GPU, with the reference's per-frame segmentation; segmenting the video saves 2--6\,min of that per take. Training on 40 takes took 5.2--5.3\,h for task~1 and 8.8\,h for task~2, whose takes have more frames.

\section{Discussion and Limitations}

\ssection{Simulation.} All results come from simulation. The simulator supplies the object masks and removes the arm by rendering without it, so the evaluation scores the motion a policy produces, not the perception of a deployed system. The rig that aligns the gripper with the recorded hand is ours, set per device by rules, and the arm stands at the same place relative to the first object picked in every layout, so reach is not tested. The success rates compare data, devices and pipeline changes; they are not predictions for a real robot.

\ssection{Scope.} One operator recorded both tasks in one room. Task~2 has one training seed per condition and no Aria counterpart, so the device's share in the second move's precision rests on the pilot. The objects' orientations are assumed rather than measured: every object must be placed square to the table edge. The inpainted images keep the arm's shadow (Fig.~\ref{fig:task2data}); blanking the image changes the policies' predictions by 5--26\,mm, so the image is used, and we did not isolate the effect of the shadow.

\ssection{Task-specific assumptions.} The rules of Section~\ref{sec:fix} were designed by inspecting failures on the same 100 layouts used to score them, with one training seed, so 93.8 of 100 is an optimistic estimate until the rules are run on fresh layouts and other tasks. Several rules use task-specific information: the reach and the approach from above act on the next object to pick after a release, and the placement rule needs the heights of the cube and the block. The placement rule also relies on an ideal re-detection of the objects, which a deployed system would have to provide by tracking them. A policy that learned to let go and to come down on an object from above would make most of the rules unnecessary; the rewritten, time-warped records go part of the way.

\section{Conclusion}

VREgo uses stereo images and on-device tracking from a consumer VR headset to train manipulation policies from human demonstrations. On the cube-on-tray task, policies trained on recordings from either device achieved simulated placements, with substantial variation across training seeds. On the two-stage task, the baseline often placed the block on the tray but deferred release or displaced the cube during approach. Adapting the training trajectories and applying task-specific execution rules increased the end-of-episode two-placement score from 2\,\% for one baseline run to a mean of 93.8\,\% over four evaluation noise seeds. The rules used simulator-derived object positions and were developed on the evaluation layouts. Independent layouts, additional training seeds, other headsets and physical robot trials are needed to assess generalisation.

\bibliographystyle{IEEEtran}
\bibliography{main.bib}

\end{document}